# Leveraging Semi-Supervised Learning to Enhance Data Mining for Image Classification under Limited Labeled Data

Aoran Shen
University of Michigan
Ann Arbor, USA

Minghao Dai
Columbia University
New York, USA

Jiacheng Hu
Tulane University
New Orleans, USA

Yingbin Liang
Northeastern University
Seattle, USA

Shiru Wang
Dartmouth College
Hanover, USA

Junliang Du*
Shanghai Jiao Tong University
Shanghai, China

*Abstract*—In the 21st-century information age, with the development of big data technology, effectively extracting valuable information from massive data has become a key issue. Traditional data mining methods are inadequate when faced with large-scale, high-dimensional and complex data. Especially when labeled data is scarce, their performance is greatly limited. This study optimizes data mining algorithms by introducing semi-supervised learning methods, aiming to improve the algorithm's ability to utilize unlabeled data, thereby achieving more accurate data analysis and pattern recognition under limited labeled data conditions. Specifically, we adopt a self-training method and combine it with a convolutional neural network (CNN) for image feature extraction and classification, and continuously improve the model prediction performance through an iterative process. The experimental results demonstrate that the proposed method significantly outperforms traditional machine learning techniques such as Support Vector Machine (SVM), XGBoost, and Multi-Layer Perceptron (MLP) on the CIFAR-10 image classification dataset. Notable improvements were observed in key performance metrics, including accuracy, recall, and F1 score. Furthermore, the robustness and noise-resistance capabilities of the semi-supervised CNN model were validated through experiments under varying noise levels, confirming its practical applicability in real-world scenarios.

## I. INTRODUCTION

In the information age of the 21st century, the speed and scale of data generation have reached unprecedented levels. With the development of big data technology, how to effectively extract valuable information from massive data has become a key issue. Traditional data mining methods are unable to cope with large-scale, high-dimensional and complex data. Especially when labeled data is scarce, the performance of traditional methods is often greatly limited. Therefore, data mining algorithms based on artificial intelligence, especially semi-supervised learning methods, have become a hot topic in current research [1].

Semi-supervised learning is an artificial intelligence technology that combines guided and unsupervised learning. It uses a small number of labeled samples and a large number of unlabeled samples to jointly train the model to enhance its generalization ability and robustness and reduce dependence on label information [2]. In actual scenarios, obtaining large-scale high-quality labeled data is often costly and cumbersome, while unlabeled data is relatively easy to obtain. Therefore, this method provides an effective way to resolve this contradiction. For example, hybrid learning has shown significant advantages in many fields such as image analysis [3-4], text classification [5-8], and risk assessment [9-13].

This study focuses on using semi-supervised learning methods to optimize data mining algorithms. By introducing semi-supervised learning, we aim to improve the algorithm's ability to effectively utilize unlabeled data, thereby achieving more accurate data analysis and pattern recognition under limited labeled data conditions. Specifically, we will adopt self-training methods to adapt to different types of data sets and application scenarios [14]. These strategies can help the model better understand the data distribution, discover potential structural information, and continuously improve the prediction performance through an iterative process.

Data mining algorithm optimization based on semi-supervised learning not only has important theoretical value but also has broad practical application prospects [15]. In the field of business intelligence, this optimization can help companies extract more insightful market trends from a large amount of user behavior data [16]; in medical research, it can accelerate the process of new drug development and find more effective treatment plans by analyzing patient genomic data [17]; Additionally, in the recommendation algorithm field [18], such optimization is crucial for enhancing the accuracy and personalization of recommendations, ensuring that the system delivers more relevant and user-specific results. By fine-tuning key parameters and employing advanced techniques, the recommendation process becomes more effective, leading to improved user satisfaction and engagement. [19], thereby enhancing user satisfaction and engagement in various applications such as e-commerce, streaming services, and social media platforms. In short, by optimizing existing algorithms based on semi-supervised learning, we hope to develop more efficient, flexible and applicable data mining tools for various practical scenarios, thereby promoting further development in related fields.

## II. RELATED WORK

IRecent advances in deep learning and semi-supervised learning have provided significant contributions to addressing challenges associated with limited labeled data. Semi-supervised learning techniques have proven effective by leveraging unlabeled data to enhance feature extraction and model robustness. For instance, the integration of self-supervised objectives with graph neural networks has demonstrated improvements in extracting meaningful patterns from complex data structures, which aligns with the need for efficient data mining and classification in this study [20]. Similarly, reinforcement learning approaches have been utilized to optimize complex systems, offering adaptive strategies that can complement semi-supervised learning models for improving classification performance [21].

The utilization of neural networks for spatiotemporal data prediction has also advanced methodologies for handling high-dimensional datasets. For example, CNN-LSTM-based frameworks have shown strong capabilities in capturing both spatial and temporal dependencies, which is relevant for enhancing feature extraction in image classification tasks [22]. Additionally, transforming multidimensional time series data into interpretable forms has introduced novel approaches for data analysis, supporting the interpretability of models in semi-supervised learning scenarios [23].

Deep learning has also played a pivotal role in improving feature representation and classification performance under various constraints. Techniques leveraging recurrent neural networks have highlighted the importance of effective time-series modeling, which can contribute to better handling of sequential data in semi-supervised learning frameworks [24]. Similarly, privacy-preserving mechanisms and data security strategies have been explored to address challenges related to ethical considerations in data-intensive tasks, offering insights that may be integrated into the development of robust semi-supervised algorithms [25].

Lastly, multimodal frameworks have showcased their potential in handling diverse datasets and enhancing model generalization. These approaches, though applied in specific contexts such as interaction design, highlight strategies that can be adapted for leveraging multiple sources of information in semi-supervised learning models [26]. Collectively, these works provide a foundation for optimizing data mining algorithms and advancing the application of semi-supervised learning in image classification tasks.

## III. METHOD

In this study, we used the autonomous training strategy in semi-supervised learning and combined it with convolutional neural networks to optimize data mining technology. Specifically, we used CNN to extract and classify image features, and gradually introduced unlabeled data through autonomous training to improve model performance as shown in Figure 1.

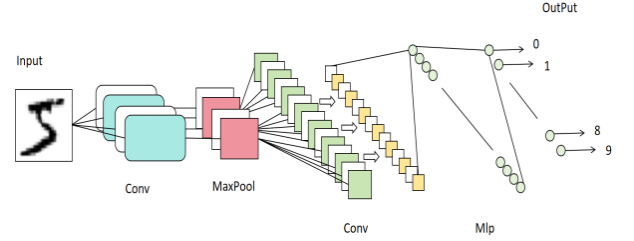

Figure 1 General architecture of convolutional neural network

First, we define an initial labeled dataset $L = \{(x_1, y_1), (x_2, y_2), ..., (x_3, y_3)\}$, where $x_i$ is the input image and $y_i$ is the corresponding label; and an unlabeled dataset $U = \{x_{l+1}, x_{l+2}, ..., x_{l+u}\}$. To ensure the quality and consistency of the data, we normalize all images so that their pixel values are normalized to the interval [0, 1]. Next, we use CNN for feature extraction. Assume that the size of the input image is $H \times W \times C$, where H and W are the height and width of the image respectively, and C is the number of channels. The output of the first convolutional layer can be expressed as:

$$O^{(1)} = \sigma(W^{(1)} * X + b^{(1)}) \quad (1)$$

$W^{(1)}$ represents the kernel weight matrix, $b^{(1)}$ denotes the bias term, and $\sigma$ represents the function of the activation. Through multi-layer convolution and pooling operations, Convolutional Neural Networks (CNNs) are capable of automatically extracting high-level features from images, allowing them to identify intricate patterns and representations crucial for tasks such as classification and object detection.

In each round of iteration, we first train the CNN model using the current labeled dataset L. Specifically, we minimize the cross entropy loss function:

$$L(\theta) = \frac{1}{L} \sum_{i=1}^{l} \sum_{c=1}^{C} y_{i,c} \log p(y_c | x_i) \quad (2)$$

Among them, $y_{i,c}$ is the true label of the i-th sample, and $p(y_c | x_i)$ is the probability distribution predicted by the model. After training, we use this model to make predictions on the unlabeled dataset U. For each unlabeled sample $x_j \in U$, calculate its classification confidence $p(y | x_j)$:

$$y_j = \arg\max p(y \mid x_j) \tag{3}$$

Select samples with confidence higher than a certain threshold $\tau$ and add them to the labeled data set L. The updated labeled data set L will be used for the next round of training. This process continues to iterate until a preset number of iterations is reached or model performance no longer improves significantly.

Through the above self-training process, we can gradually expand the labeled data set, thereby improving the generalization ability of the model. In addition, the feature maps extracted by CNN at each layer can also be used for further data mining tasks. We use t-SNE (T-distributed Stochastic Neighbor Embedding) to map high-dimensional features to two-dimensional space for visual analysis and to discover clustering structures and anomalies in the data [27]. In addition, by analyzing features at different levels, we can identify which features have an important impact on the final classification results. This can be achieved by calculating the importance score of the feature [28], and we use the gradient-weighted class activation mapping (Grad-CAM) method:

$$CAM = \sum_k w_k \cdot A^k \tag{4}$$

In this way, we can locate the key areas in the image. In summary, by combining the self-training method of semi-supervised learning and the CNN of deep learning, we not only improve the accuracy of image classification, but also gain deeper insights in data mining. This integrated approach provides a new way to deal with large-scale and complex data sets.

## IV. EXPERIMENT

### A. Datasets

In accordance with the principles of semi-supervised learning, the CIFAR-10 dataset is partitioned into two distinct subsets: a labeled dataset and an unlabeled dataset. The CIFAR-10 dataset, a standard benchmark for image classification tasks, comprises 60,000 color images evenly distributed across 10 categories, with each image having dimensions of 32×32 pixels. To construct the labeled dataset (L), 1,000 images are randomly selected from each category, resulting in a total of 10,000 labeled images that serve as the primary data for supervised training. The remaining 50,000 images, devoid of labels, constitute the unlabeled dataset (U). This partitioning strategy reflects a realistic imbalance between labeled and unlabeled data, capturing the challenges of real-world scenarios where labeled data is often scarce due to the high cost and time required for annotation, whereas unlabeled data is more readily accessible in large volumes.

The rationale behind this division is to simulate and evaluate a realistic use case where only a limited amount of labeled data is available for training, necessitating the use of semi-supervised learning techniques to maximize performance. This experimental setup allows us to assess the ability of convolutional neural network (CNN) models to effectively learn and generalize under conditions of scarce labeled data.

By leveraging the abundance of unlabeled images in U, the model can explore semi-supervised methods to improve its performance. This mirrors practical applications across various domains, such as healthcare, autonomous driving, and natural language processing, where annotating data is labor-intensive or requires specialized expertise.

Moreover, this dataset split creates an ideal testing ground for evaluating the capability of semi-supervised learning frameworks to balance the trade-off between limited supervision and the exploitation of vast unlabeled data. Through this approach, we aim to provide a comprehensive understanding of how semi-supervised CNN models operate in scenarios that closely mimic real-world challenges, demonstrating their potential effectiveness and robustness in practical applications.

### B. Experimental Results

To thoroughly assess the effectiveness and superiority of the convolutional neural network (CNN) model built on a semi-supervised learning framework for data mining tasks, we devised a comprehensive series of experiments. The primary goal of these experiments is to measure how well the model performs in image classification tasks using the CIFAR-10 dataset, particularly under the constraints of limited labeled data. Our approach involves leveraging a self-training methodology to progressively expand the labeled dataset and evaluate how the model's performance evolves over time.

The self-training process begins with the initially labeled dataset, and during each iteration, the model is tasked with making predictions on the unlabeled dataset. From these predictions, we identify the unlabeled samples for which the model demonstrates the highest confidence scores—indicating that it is most certain about the class assignments of these samples. These high-confidence samples are then moved from the unlabeled dataset to the labeled dataset, effectively augmenting the amount of labeled data available for training. The model is subsequently retrained using the updated labeled dataset. This iterative process continues until one of two conditions is met: either a predefined number of iterations is reached, or the model's performance improvements plateau, meaning no significant gains are observed in subsequent iterations. To further demonstrate the advantages of the proposed method, we compare the CNN model based on semi-supervised learning with traditional machine learning methods, including multi-layer perceptron (MLP) [29], support vector machine (SVM) and XGBoost [30]. For these traditional methods, we will use the same initial labeled data set L, and unlabeled data set U, and adopt the standard supervised learning process for training. We will then evaluate the performance of all models on the test set, comparing their performance under different proportions of labeled data. In this way, we can intuitively see how CNN models based on semi-supervised learning can outperform other methods with limited labeled data.

In addition to the comparison of classification performance, we also focus on the generalization ability and robustness of the model. To do this, we will conduct a cross-validation experiment, dividing the dataset into multiple subsets, ensuring that each subset contains a certain proportion of labeled and unlabeled data. Through multiple cross-validation, we are able to reliably assess the stability and consistency of the model under different data partitions. In addition, we will also analyze the performance of the model when processing noisy data, and observe the robustness and anti-interference ability of the model by introducing different levels of noise into unlabeled data. These experiments will help us comprehensively understand the advantages and limitations of semi-supervised learning-based CNN models in practical applications. We first show the loss function drop graph during training in Figure 2.

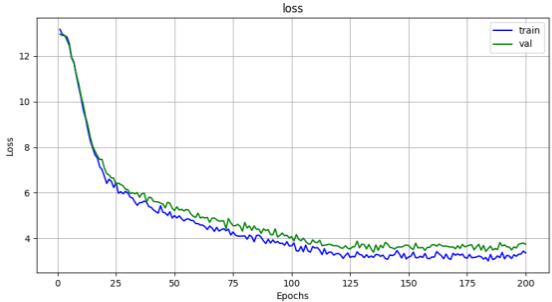

Figure 2 Loss function drop graph

We then present the corresponding experimental results, which are summarized in Table 1.

Table 1 The Experiment Result

| Setting | Acc | RECALL | F1 |
| --- | --- | --- | --- |
| SVM | 0.579 | 0.561 | 0.563 |
| XGBOOST | 0.721 | 0.723 | 0.722 |
| MLP | 0.833 | 0.842 | 0.831 |
| Ours | 0.897 | 0.901 | 0.899 |

It clearly indicates that the CNN model based on semi-supervised learning performs significantly better than the traditional machine learning method on the CIFAR-10 data set. Specifically, our model achieved 0.897, 0.901 and 0.899 in accuracy (Acc), recall (RECALL) and F1 score respectively, which is significantly higher than the performance of SVM, XGBoost and MLP. This result shows that by combining self-training methods and deep learning techniques, we can more effectively utilize limited labeled data and large amounts of unlabeled data to improve the model's classification performance. Especially compared with SVM, our model improves accuracy by nearly 32%, which illustrates the advantages of deep learning methods in processing complex image features.

Further analysis of the specific indicators of each model shows that SVM performs the worst on all evaluation indicators, with an accuracy of only 0.579, a recall of 0.561, and an F1 score of 0.563. This may be due to the limitations of SVM in processing high-dimensional, non-linear data. Especially when facing large-scale image data, SVM has difficulty in effectively capturing the subtle features in the image. In comparison, the performance of XGBoost and MLP is much better. These results show that tree model-based XGBoost and neural network-based MLP have stronger capabilities in handling image classification tasks, especially in feature extraction and pattern recognition.

However, even between XGBoost and MLP, our semi-supervised learning-based CNN model still performs well. Compared to the Multi-Layer Perceptron (MLP), our model achieves significant improvements, with accuracy increasing by approximately 6.4 percentage points, recall by approximately 5.9 percentage points, and F1 score by approximately 6.8 percentage points. This significant improvement verifies the superiority of deep learning methods in image classification tasks. Through the self-training process, our model is able to gradually increase labeled data in each iteration, thereby continuously optimizing model parameters and improving generalization ability and robustness. In addition, the CNN model can automatically learn multi-level image features, which is particularly important for processing complex visual tasks.

In summary, the experimental results clearly demonstrate the superior performance of the CNN model based on semi-supervised learning in image classification tasks. This model not only exceeds traditional machine learning methods in metrics such as accuracy, recall, and F1 score but also exhibits greater practicality and scalability for real-world applications. By fully utilizing unlabeled data, our method is able to maintain high classification performance under resource constraints, which is very valuable for many real-world application scenarios as shown in Figure 3.

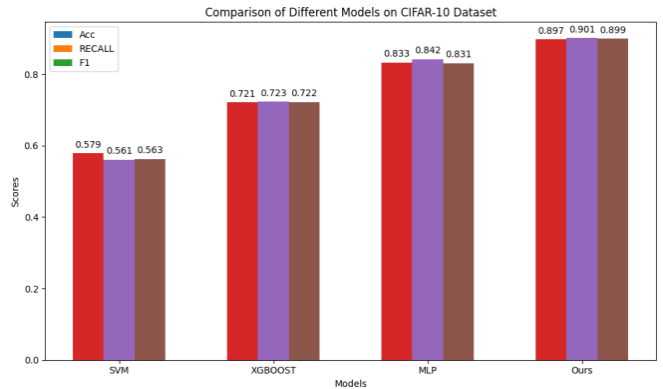

Figure 3 Experimental results bar graph

V. CONCLUSION

This study leverages semi-supervised learning to enhance data mining algorithms, especially in scenarios with limited labeled samples, by utilizing unlabeled data to boost the accuracy of data analysis and pattern recognition. Using an autonomous training mechanism integrated with a convolutional neural network (CNN), the approach enables efficient image feature extraction and classification. The CNN is iteratively tuned with the labeled set *L* to minimize cross-entropy loss and gradually incorporates unlabeled samples to expand the labeled dataset and improve the model's generalization. The study also employs t-SNE for dimensionality reduction of high-dimensional CNN feature maps, offering a visual understanding of clustering structures and outliers in a two-dimensional space. Additionally, Grad-

CAM is used to calculate feature importance scores, allowing for the identification of key regions in images and better interpretation of model decisions. This comprehensive approach not only improves image classification accuracy but also introduces new methodologies for analyzing large-scale, complex datasets. Testing on the CIFAR-10 dataset, with its 60,000 32x32-pixel images across 10 categories, demonstrates the efficiency and accuracy of the proposed method under limited labeled data conditions. This semi-supervised optimization of data mining algorithms holds significant theoretical value and offers wide-ranging practical applications. For instance, in business intelligence, it can help extract market trends from extensive user behavior data; in medical research, it may expedite drug discovery and optimize treatment strategies through genomic data analysis; and in social sciences, it can uncover underlying patterns in human behavior and societal phenomena.

In summary, by refining data mining algorithms with semi-supervised learning, this study aims to develop more efficient, adaptable tools for a variety of applications, advancing both academic research and practical innovations in data-intensive fields.